\documentclass[10pt,twocolumn,letterpaper]{article}

\usepackage{iccv}              

\usepackage{multirow}

\usepackage[dvipsnames]{xcolor}
\usepackage{calc}

\newcommand{\tocite}[1]{{\bf\color{red}[TOCITE]}}

\newcommand{\PAR}[1]{\vskip4pt \noindent{\bf #1~}}

\newcommand{\sysname}[0]{Diffuman4D}

\definecolor{tabfirst}{rgb}{1, 0.7, 0.7}
\definecolor{tabsecond}{rgb}{1, 0.85, 0.7}
\definecolor{tabthird}{rgb}{1, 1, 0.7}
\definecolor{tabgray}{rgb}{0.9, 0.9, 0.9}
\definecolor{turquoise}{cmyk}{0.65,0,0.1,0.3}
\definecolor{purple}{rgb}{0.65,0,0.65}
\definecolor{darkgreen}{rgb}{0, 0.5, 0}
\definecolor{orange}{rgb}{0.8, 0.6, 0.2}
\definecolor{red}{rgb}{0.8, 0.2, 0.2}
\definecolor{darkred}{rgb}{0.6, 0.1, 0.05}
\definecolor{blueish}{rgb}{0.0, 0.3, .6}
\definecolor{light_gray}{rgb}{0.7, 0.7, .7}
\definecolor{pink}{rgb}{1, 0, 1}
\definecolor{greyblue}{rgb}{0.25, 0.25, 1}
\definecolor{titlepurple}{RGB}{66,93,205}
\definecolor{titleblue}{RGB}{21,129,203}
\definecolor{titlegreen}{RGB}{57,145,136}

\newcommand\rgt{\aftergroup\mathclose\aftergroup{\aftergroup}\right}

\definecolor{iccvblue}{rgb}{0.21,0.49,0.74}
\usepackage[pagebackref,breaklinks,colorlinks,allcolors=iccvblue]{hyperref}
\usepackage{enumitem}
\usepackage{float}
\usepackage{multirow}

\def\paperID{4709} 
\def\confName{ICCV}
\def\confYear{2025}

\title{\textcolor{titlepurple}{Diffu}\textcolor{titleblue}{man}\textcolor{titlegreen}{4D}: 4D Consistent Human View Synthesis from \\Sparse-View Videos with Spatio-Temporal Diffusion Models}

\author{
Yudong Jin$^1$
\quad
Sida Peng$^1$
\quad
Xuan Wang$^2$
\quad
Tao Xie$^1$
\quad
Zhen Xu$^1$\\
\quad
Yifan Yang$^1$
\quad
Yujun Shen$^2$
\quad
Hujun Bao$^1$
\quad
Xiaowei Zhou$^1$$^{\dagger}$
\vspace{2mm}
\\
\centerline{$^1$Zhejiang University \quad $^2$Ant Research}
}

\newcommand\blfootnote[1]{
  \begingroup
    \renewcommand\thefootnote{}
    \footnote{#1}
    \addtocounter{footnote}{-1}
  \endgroup
}

\begin{document}

\twocolumn[{
    \renewcommand\twocolumn[1][]{#1}
    \vspace{-3em}
    \maketitle
    \begin{center}
        \captionsetup{type=figure}
        \centering
        \includegraphics[width=1.0\textwidth]{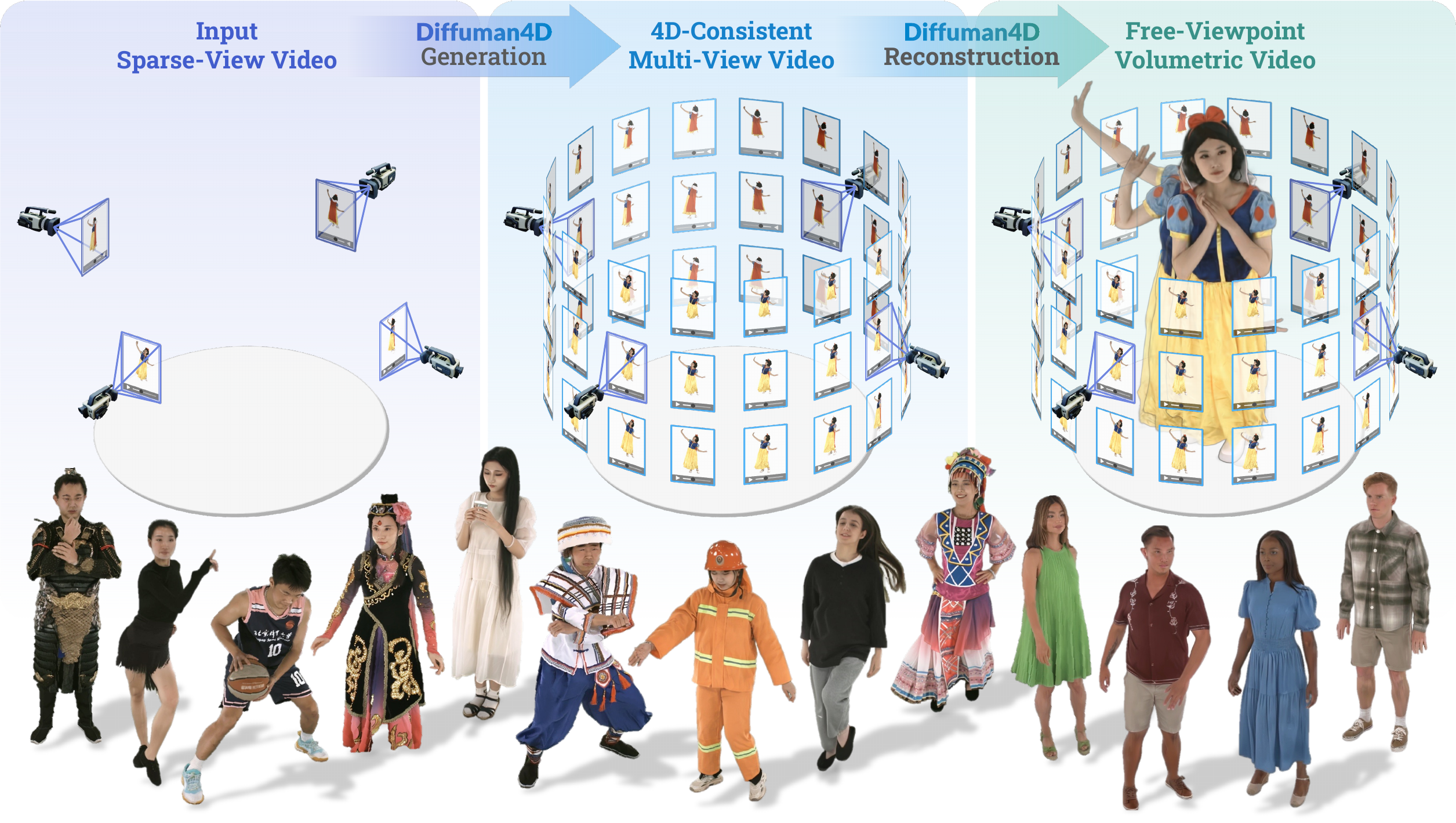}
        \vspace{-0.2in}
        \caption{\sysname{} enables high-fidelity free-viewpoint rendering of human performances from sparse-view videos. The bottom row shows representative results.}
        \vspace{0in}
        \label{fig:teaser}
    \end{center}
}]

\blfootnote{$\dagger$ Corresponding author: Xiaowei Zhou.}

\begin{abstract}
This paper addresses the challenge of high-fidelity view synthesis of humans with sparse-view videos as input.
Previous methods solve the issue of insufficient observation by leveraging 4D diffusion models to generate videos at novel viewpoints.
However, the generated videos from these models often lack spatio-temporal consistency, thus degrading view synthesis quality.
In this paper, we propose a novel sliding iterative denoising process to enhance the spatio-temporal consistency of the 4D diffusion model.
Specifically, we define a latent grid in which each latent encodes the image, camera pose, and human pose for a certain viewpoint and timestamp, then alternately denoising the latent grid along spatial and temporal dimensions with a sliding window, and finally decode the videos at target viewpoints from the corresponding denoised latents. 
Through the iterative sliding, information flows sufficiently across the latent grid, allowing the diffusion model to obtain a large receptive field and thus enhance the 4D consistency of the output, while making the GPU memory consumption affordable.
The experiments on the DNA-Rendering and ActorsHQ datasets demonstrate that our method is able to synthesize high-quality and consistent novel-view videos and significantly outperforms the existing approaches.
See our project page for interactive demos and video results: \href{https://diffuman4d.github.io/}{https://diffuman4d.github.io/}.
\end{abstract}

\section{Introduction}
\label{sec:intro}
This paper aims to address the problem of high-quality 4D view synthesis of humans in motion from sparse-view videos, which has wide applications in augmented reality, film production, sports broadcasting, etc.
Traditional multi-view stereo-based methods~\cite{schoenberger2016sfm,schoenberger2016mvs,Galliani_2015_ICCV} and recent neural rendering-based methods~\cite{yang2023gs4d,Wu_2024_CVPR,xu20244k4d} generally require a dense array of synchronized cameras to capture a sufficient number of views for high-quality reconstruction, making them difficult to apply to real-world scenarios.
When input views become sparse, these methods tend to fail as the insufficient observations make the reconstruction problem ill-posed.

An intuitive solution to this problem is to leverage conditional image or video generative models to generate novel views conditioned on the input views ~\cite{wu2024cat4d,xie2024sv4d,watson2024controlling}.
By leveraging the attention mechanism, these methods inject spatial and temporal control signals into video generative models, aiming to produce human images at target viewpoints and timestamps.
However, these methods often struggle with the spatio-temporal consistency of generated images, especially when the human topology and cloth deformations are complex.
A key reason is that, due to GPU memory limitations, these methods often have to generate the target images in multiple passes, and the inherent probabilistic nature of generative models leads to variance among the outputs.

In this paper, we propose a novel spatio-temporal diffusion model for generating 4D consistent multi-view human videos.
Our key innovation lies in a novel sliding iterative denoising process that ensures the 4D consistency of our model's outputs.
Specifically, given a sparse-view video, we first encode image observations and camera parameters as conditioning latents, and define noise latents at target viewpoints and timestamps, forming a 4D latent grid.
To denoise the latent grid, we use a window that alternately slides forward and backward in the spatial and temporal dimensions.
In contrast to the previous method~\cite{wu2024cat4d} that executes the full denoising process at each sliding operation, our model only performs a few denoising steps during sliding.
This iterative sliding strategy ensures sufficient information propagation across the latent grid, enabling the diffusion model to incorporate surrounding 4D signals to produce each target output, while dynamically adjusting their influence based on spatio-temporal distance.

To further boost the consistency of our generative model, we exploit the 3D human skeleton sequence as a structural prior to guide the generation process.
Concretely, our method first extracts a 3D human skeleton sequence from the given sparse-view video using an off-the-shelf 3D pose estimator.
Then, we project the skeletons into the image space at each viewpoint and each timestamp, serving as the conditioning signal alongside camera parameters and image observations to guide the spatio-temporal diffusion model. 
Finally, we decode the videos at target viewpoints from the corresponding denoised latents, and reconstruct high-quality 4D Gaussian Splatting (4DGS) \cite{xu2024longvolcap,yang2023gs4d,Wu_2024_CVPR} based on the input-view and the synthesized novel-view videos. 

To enable model training, we meticulously process the DNA-Rendering~\cite{2023dnarendering} dataset by recalibrating camera parameters, optimizing image color correction matrices (CCMs), predicting foreground masks, and estimating human skeletons. We compare our method against state-of-the-art methods on DNA-Rendering and ActorsHQ datasets, highlighting the performance of our framework in capturing detailed human motion and appearance from sparse-view inputs. 

In summary, our contributions are as follows:
\begin{itemize}[leftmargin=2em]
    \item We introduce Diffuman4D, a novel diffusion model that generates spatio-temporally consistent and high-resolution (1024p) human videos from sparse-view video inputs.
    \item We propose a sliding iterative denoising mechanism that enhances both spatial and temporal consistency of generated long-term videos while maintaining efficient inference.
    \item We design a human pose conditioning scheme to enhance appearance quality and motion accuracy of generated human videos.
    \item We plan to release our processed version of the DNA-Rendering dataset, which we believe will benefit future research in this area.
\end{itemize}

\section{Related Work}
\label{sec:related_work}

\PAR{4D reconstruction from dense views.}
Reconstructing dynamic 3D human performances from captured videos and performing novel view synthesis to create immersive playbacks has been a long-standing research problem in computer vision and graphics.
Traditional methods tend to leverage complicated hardware, such as dense camera arrays \cite{starck2007surface, collet2015high, kanade1997virtualized, grau2003studio, starck2005virtual} or depth sensors \cite{aitpayev2012creation, shapiro2014rapid, tong2012scanning, bogo2015detailed, newcombe2015dynamicfusion}, to reconstruct high-fidelity human performances.
Recent advancements in neural scene representation, namely neural radiance field (NeRF)~\cite{mildenhall2021nerf} and 3D Gaussian Splatting (3DGS)~\cite{kerbl3Dgaussians}, have demonstrated remarkable success in static 3D scene reconstruction.
\cite{li2022neural,wang2022fourier,fridovich2023k,cao2023hexplane,shao2023tensor4d,yang2023gs4d,duan20244d,xu2024longvolcap,xu20244k4d} propose to lift this base 3D representation (NeRF or 3DGS) to 4D by adding an additional temporal dimension, enabling it to model temporal variations in dynamic scenes.
However, these methods still heavily rely on densely distributed and well-synchronized multi-view video inputs and exhibit severe overfitting under sparse-view conditions, which greatly limits their accessibility.

\PAR{4D reconstruction from sparse views.}
To alleviate the requirement of dense multi-view inputs, some methods~\cite{peng2021neural,weng2022humannerf} leverage human priors such as SMPL~\cite{loper2023smpl} to guide the reconstruction process.
\cite{peng2021animatable,hu2023gauhuman,xu2024relightable} propose constructing a 3D static canonical space with a neural field and learning a deformation field~\cite{pumarola2020d,fang2022fast,park2021nerfies,park2021hypernerf,li2024animatablegaussians} based on the SMPL human prior to map dynamic elements back to this canonical space.
While effective in certain scenarios, these approaches face challenges in accurately estimating shape deformations, particularly when dealing with complex outfits or rapid motions.
On the other hand, methods like~\cite{lin2022enerf,chen2024mvsplat,charatan23pixelsplat,liu2024mvsgaussian,zheng2024gpsgaussian} explore the use of depth priors, such as stereo~\cite{lipson2021raft} or multi-view~\cite{yao2018mvsnet,yao2019recurrent} depth estimation for generalizable scene reconstruction and novel view synthesis.
However, these methods are highly dependent on the accuracy of depth estimation, often struggling in cases involving occlusions, textureless regions, or extremely sparse viewpoints.

\PAR{4D generation.}
Recent developments in 3D content generation~\cite{poole2022dreamfusion,shi2023mvdream,wang2023imagedream,hollein2024viewdiff,liu2023zero1to3,liu2023syncdreamer,gao2024cat3d} and video diffusion technologies~\cite{brooks2024video,hong2022cogvideo,yang2024cogvideox,kong2024hunyuanvideo} provide a promising direction for handling challenging scenarios mentioned above by introducing generative data priors to the reconstruction pipeline.
\cite{singer2023text,bahmani2024tc4d,bahmani20244d,ren2023dreamgaussian4d,yin20234dgen,zhao2023animate124,zheng2024unified,zhang20254diffusion} leverage the Score Distillation Sampling (SDS)~\cite{poole2022dreamfusion} to extract 4D representations from image or video diffusion models. However, SDS limits their scalability to large-scale 4D reconstruction tasks due to its high computational cost and tends to produce over-smoothed geometries and unrealistic textures.
To overcome these limitations, \cite{li2025vivid,pan2024efficient4d,yang2024diffusion,zeng2024stag4d,xie2024sv4d} propose to condition the diffusion models to generate spatio-temporal consistent multi-view videos, which can be further used for 4D reconstruction. However, these methods only focus on object-level generation.
Recently, CAT4D~\cite{wu2024cat4d} has made a step forward by proposing a multi-view consistent video diffusion model that can handle general scenes by leveraging general time embedding and Plücker embedding~\cite{jia2020plucker} for temporal and spatial consistency. Despite the improvement, CAT4D still struggles with complex human generation due to the heavy shape distortion and self-occlusions caused by the motion of soft structures like hair and clothes, and it is difficult for diffusion models to solve these ill-posed problems by only relying on the general conditions. In this paper, we propose to introduce additional human-specific priors to address these challenges.

\section{Method}
\label{sec:method}

We reconstruct human performances from sparse-view videos in two stages.  
First, we transform the input sparse-view videos into dense multi-view videos using our spatio-temporal diffusion model. Then, we reconstruct the human performance by optimizing a 4D Gaussian Splatting (4DGS) from these generated multi-view videos. 
We first describe our spatio-temporal diffusion model (\cref{sec:model}) and the denoising mechanism for generating spatio-temporal consistent multi-view videos (\cref{sec:denoising}).
Then, we describe the skeleton conditioning scheme (\cref{sec:skeleton}), and the method we used to reconstruct human performances (\cref{sec:reconstruction}).

\subsection{Spatio-Temporal Diffusion Model}
\label{sec:model}

\begin{figure*}[ht!]
    \centering
    \includegraphics[width=0.99\linewidth]{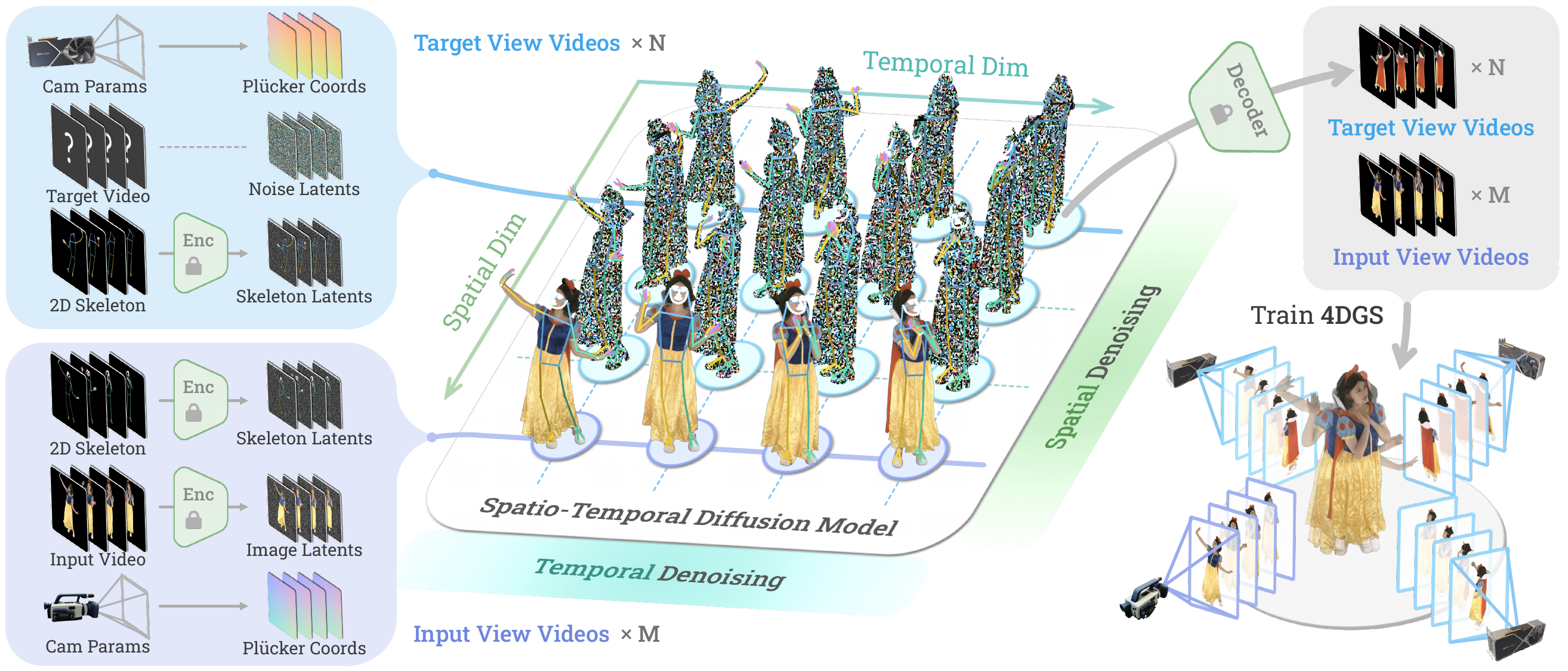}
    \vspace{-0.1in}
    \caption{\textbf{Overview of \sysname{}}. 
Our model takes $M$ input-view videos as input, generates $N$ target-view videos, and reconstructs a 4DGS of the human using both input and generated videos. 
Specifically, the input-view videos are encoded into the latent space using a pretrained VAE.
The 3D human‐skeleton sequence is projected into each view, rendered as RGB maps, and encoded into the same latent space.
In addition, the camera parameters are encoded as Plücker coordinates~\cite{zhang2024cameras}.
The skeleton latents and Plücker coordinates are then concatenated with the image latents at input views or the noise latents at target views, forming input samples or target samples, respectively.
The samples across all views and timestamps form a sample grid and are fed into our spatio-temporal diffusion model that denoises the samples using a sliding iterative denoising mechanism (see \cref{sec:denoising}).
Then, the denoised image latents for the target views are decoded into target-view videos using a pretrained VAE decoder.
Finally, the 4DGS is reconstructed using an off-the-shelf approach~\cite{xu2024longvolcap} (see \cref{sec:reconstruction}), enabling real-time novel view rendering.}
    \vspace{-0.1in}
    \label{fig:pipeline}
\end{figure*}

\PAR{Pipeline.}
As illustrated in \cref{fig:pipeline}, our spatio-temporal diffusion model takes $M$-view videos as input and aims to generate $N$-view target videos, where all videos consist of $T$ frames. We first encode the input videos into a latent space using a pretrained VAE and generate noise latents for the target videos. These latents are structured into a sample grid of size $(N+M) \times T$, where the two axes represent the spatial (multi-view) dimension and the temporal (video) dimension. Each sample within the grid comprises an input image latent (or a target noise latent) along with its corresponding conditioning embeddings, which consist of a skeleton latent and Plücker coordinates (see \cref{sec:skeleton} for details).
We then employ a sliding iterative approach to progressively denoise the sample grid (see \cref{sec:denoising} for details). Next, we decode the target image latents into corresponding videos. Finally, we reconstruct a high-quality 4D Gaussian Splatting (4DGS) representation of the human performance using both the input-view and target-view videos, enabling real-time rendering.

\PAR{Architecture.}
Our model follows the architecture of multi-view latent diffusion models~\cite{gao2024cat3d,shi2023mvdream}.
Specifically, the model employs 3D self-attention layers to enable information exchange across images. The images are encoded into latent representations through a pretrained VAE, allowing the model to learn the joint distribution in the latent space.
We disable the text conditioning by setting the input prompt to an empty string.

\subsection{Sliding Iterative Denoising}
\label{sec:denoising}

To achieve high-quality 4D human reconstruction, the input data must be sufficiently dense in both spatial (multi-view) and temporal (video) dimensions. 
For instance, reconstructing a 10-second 4D human typically requires tens of thousands of input images. 
Due to GPU memory constraints, existing video diffusion models can only denoise a limited number of images per inference, requiring the entire image sequence to be split into hundreds of groups. 
Since these denoising iterations operate independently, the generated images exhibit inconsistency due to variations in the diffusion denoising process.
The recent approach~\cite{wu2024cat4d} introduces a sliding-window strategy that separately complete the denoising process within each sliding window and then apply median filtering to the overlapping samples, thereby reducing the variance of diffusion denoising. 
However, this method still suffers from inconsistency when generating long-sequence samples, and performing multiple denoising iterations significantly increases the inference time.
See \cref{fig:ablation_denoising} for comparative results.

\PAR{Sliding iterative denoising.}
To address the aforementioned challenges, we propose a sliding iterative denoising mechanism that enhances consistency in long-sequence generation by leveraging rich contextual information during the denoising process. 
Specifically, given a target sample sequence, we set a context window of length $W$ that slides over the sequence with a fixed stride $S$. 
In each iteration, we concatenate the target samples with input samples and feed them into the diffusion model for denoising $P$ steps. 

\begin{figure*}[t!]
    \centering
    \begin{subfigure}[b]{0.39\linewidth}
        \centering
        \includegraphics[width=\linewidth]{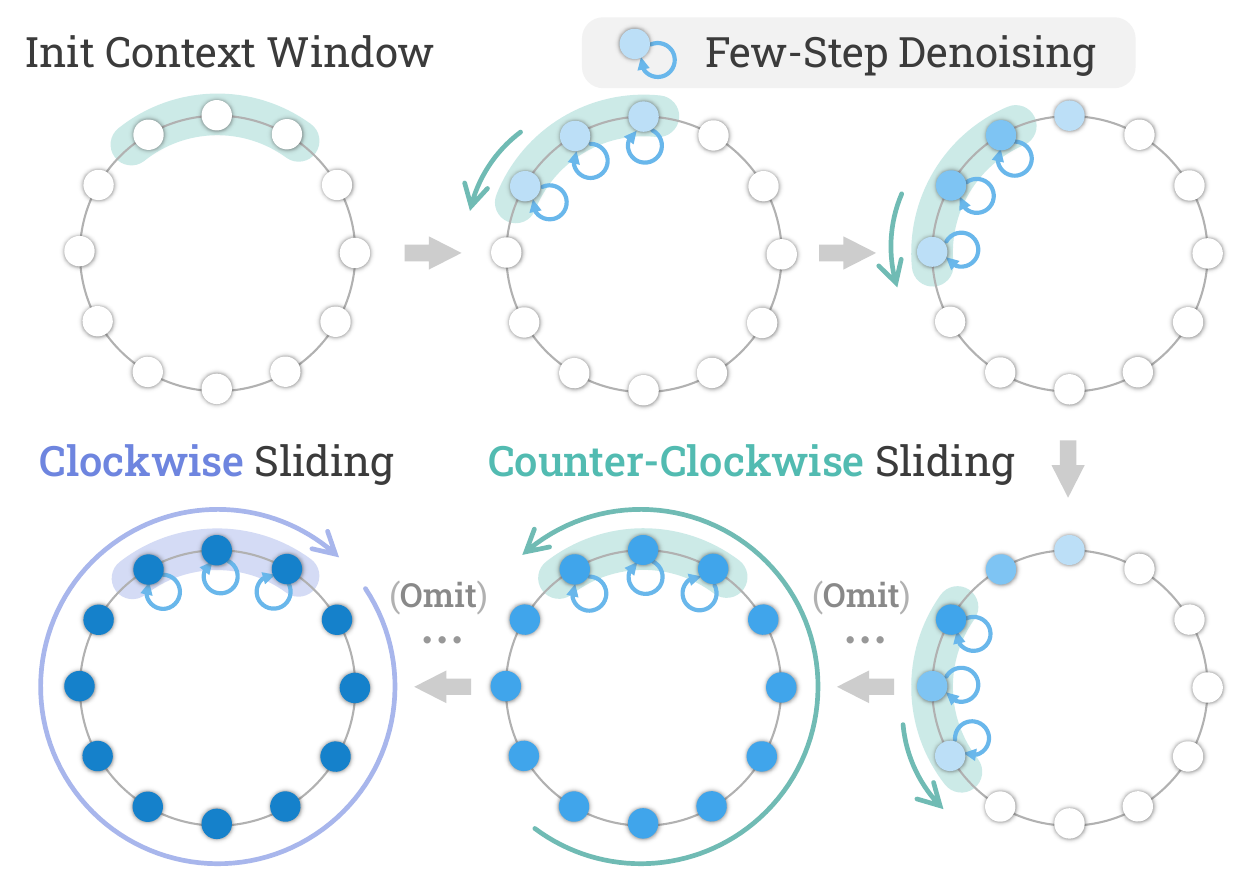}
        \caption{\textbf{Sliding iterative denoising mechanism}}
        \label{fig:method_context_denoising}
    \end{subfigure}
    \hfill
    \begin{subfigure}[b]{0.57\linewidth}
        \centering
        \includegraphics[width=\linewidth]{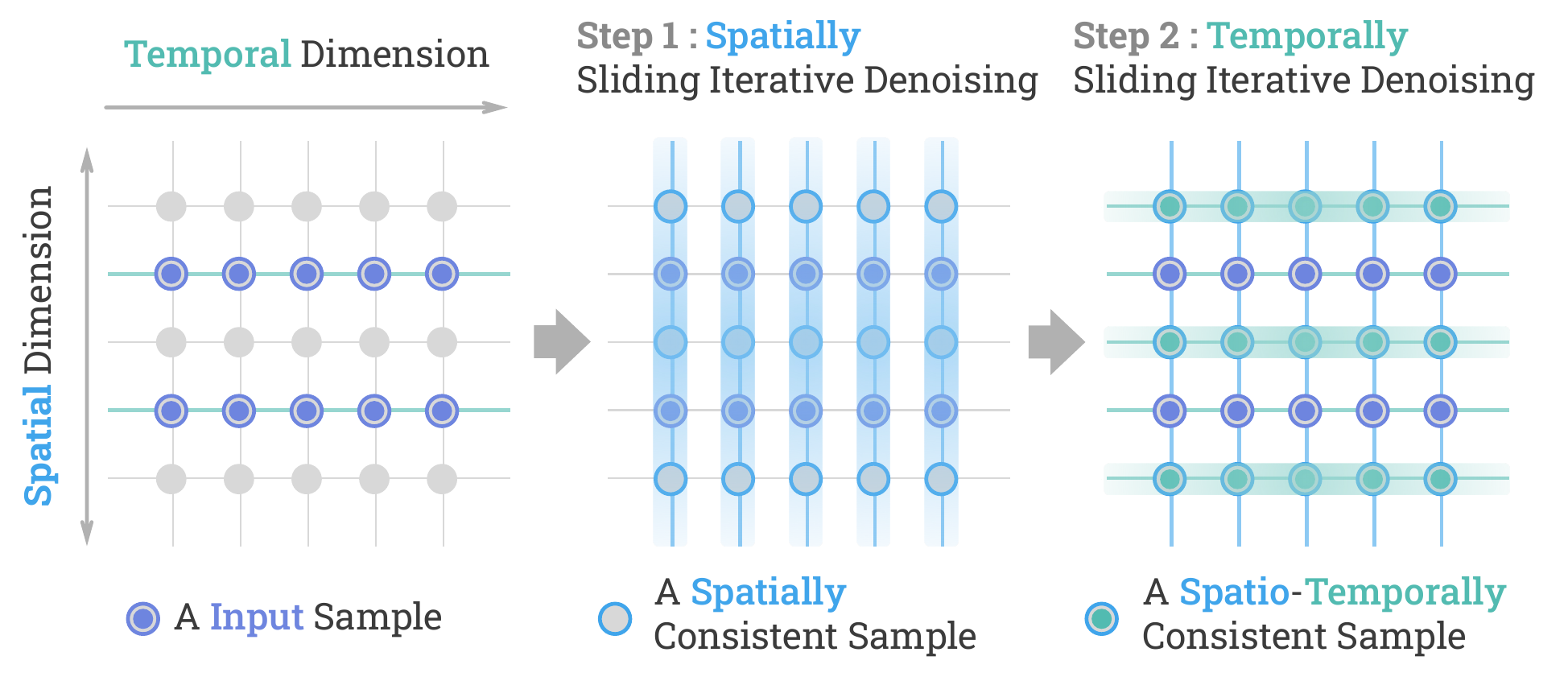}
        \caption{\textbf{Spatio-temporal denoising process}}
        \label{fig:method_denoising_process}
    \end{subfigure}
    \vspace{-0.1in}
    \caption{\textbf{Illustration of our denoising framework.} (a) \textbf{Sliding iterative denoising mechanism}. Given a view sequence arranged in a circular manner, we initialize a context window of length $W$ (here, $W=3$). The window slides counterclockwise with a stride of $S$ (here, $S=1$), and in each iteration, the samples within the context window are fed into the model for $P$ denoising steps. After completing a full rotation, we reverse the direction of the window, allowing it to slide clockwise for another full cycle. Consequently, each sample undergoes a total of $D=2PW/S$ denoising steps. (b) \textbf{Spatio-temporal denoising process}. Given an $M$-view, $N$-frame video (here, $M = 2$, $N = 5$) and diffusion inference steps $D$, the model first performs denoising in the spatial dimension for $D / 2$ steps, producing spatially consistent samples (latents). It then performs denoising in the temporal dimension for the remaining $D / 2$ steps, resulting in spatio-temporally consistent samples. Sliding iterative denoising is employed in both spatial and temporal denoising processes. Each column or row can be denoised in parallel.
    }
    \vspace{-0.1in}
    \label{fig:method_denoising_combined}
\end{figure*}

As illustrated in \cref{fig:method_context_denoising}, given a human-centric sample sequence (e.g., camera views arranged circularly), we first slide the context window counter-clockwise to enable information propagation along that direction. 
Subsequently, we reverse the sliding direction to clockwise, allowing bidirectional context aggregation. 
Formally, each sample is denoised $D = 2 \times P \times W/S$ steps in total. 
We accordingly set the diffusion inference steps to $D$, therefore the generation is completed after the above process. 

The same operation can be applied to the temporal sample sequence, enabling each sample to aggregate both past and future contexts during denoising. 
It is worth noting that multiple sample sequences can be denoised in parallel using multiple GPUs.
The sliding iterative denoising mechanism allow the diffusion model to harmonize the consistency of the target samples in both spatial and temporal dimensions, enabling the generation of high-quality 4D image grids.

\PAR{Alternating denoising.} 
Following the previous work~\cite{wu2024cat4d}, we adopt an alternating denoising strategy to further improve the spatial and temporal consistency of generated images. \cref{fig:method_denoising_process} illustrates our denoising process. 
Given an $M$-view, $T$-frame video, we first denoise target samples in the spatial dimension for $D/2$ steps, conditioning on the $M$ input views at the corresponding time, and then perform denoising in the temporal dimension for the remaining $D/2$ steps, conditioning on the $W$ frames within the corresponding time range from the nearest-distance view. 
This ensures the generation of spatio-temporally consistent samples.

By combining this strategy with sliding iterative denoising, each sample perceives surrounding spatial information from adjacent camera views and temporal information from neighboring video frames through sliding context windows. 
Furthermore, each target sample acts as a local center, and samples closer to it undergo more joint denoising steps within the context window. 
This mechanism aligns with the nature of 4D data: closer samples exhibit stronger correlations, which require more intensive consistency constraints.

\subsection{Skeleton-Conditioned Diffusion}
\label{sec:skeleton}

Previous 4D generation methods~\cite{wu2024cat4d,watson2024controlling} were designed for general scenes by directly injecting spatial signals (camera embeddings) and temporal signals (timestamp embeddings) into diffusion models. 
However, these methods face significant challenges when generating spatio-temporally consistent human images.
First, while Plücker coordinates provide pixel-wise camera pose signals, the generated images often exhibit noticeable pose errors from the input cameras.
Second, human motion often involves substantial deformation (such as hair and loose clothing), leading to various shape distortions and self-occlusions. 
Diffusion models struggle to stabilize the generation process.

\PAR{Skeleton conditioning.}
To address these challenges, we introduce additional human-specific conditioning signals to constrain the model’s generation space, leading to more accurate and consistent human image synthesis. 
Considering the vast diversity of human data (such as different clothing, body shapes, and genders), we require an intermediate human representation that ensures precise conditioning signals for the model. 
3D human skeleton sequence emerges as an ideal choice to meet these requirements, as it can be easily recovered from sparse-view videos and provides a consistent 4D human structure.
These 3D skeletons can be projected into 2D space, capturing a snapshot from a specific camera at a given timestamp, to enhance both temporal and spatial conditioning signals for diffusion models.

Specifically, we first estimate 2D human skeletons using Sapiens~\cite{khirodkar2024sapiens} and triangulate them to obtain a 3D skeleton sequence. 
We then project these skeletons into each view and render them as an RGB map, assigning different colors to different body parts to enrich the conditioning information.
This RGB map is then encoded into the latent space using a pretrained VAE, serving as a pixel-aligned feature that significantly enhances generation quality for complex human poses.

\PAR{Skeleton-Plücker mixed conditioning.}
However, skeleton predictions are often incomplete for individuals wearing complex clothing, leading to a degradation in the completeness of pose control signals.
Additionally, since skeleton representations do not contain explicit occlusion information, they introduce inherent ambiguities of front-back symmetry. See \cref{fig:ablation_conditioning} for an example. 
To mitigate these limitations, we retain Plücker coordinate conditioning to provide explicit camera pose information, thereby improving the robustness of the generation process.

\subsection{4DGS Reconstruction}
\label{sec:reconstruction}

Based on the approach above, our model can generate dense-view videos with spatio-temporal consistency. The output videos can be fed into any existing 4D reconstruction pipeline to yield a 4D human representation. 
In our experiments, we employ LongVolcap~\cite{xu2024longvolcap,xu2023easyvolcap} as our reconstruction method, which is an enhanced version of 4DGS that can effectively reconstruct a long volumetric video with a temporally hierarchical Gaussian representation.

\section{Experiments}
\label{sec:experiments}

\subsection{Implementation Details}

We train the model on either spatial sample sequences or temporal sample sequences, each with a total length of $M + N = 16$.
For the spatial sample sequence, all samples are captured simultaneously from different cameras, and the model is trained to generate $N=12$ target samples from $M=4$ conditional samples.
For the temporal sample sequence, the model is trained to generate $N=8$ consecutive samples from a target camera $C_T$, conditioned on $M=8$ samples over the same time range from a reference camera $C_{R}$, which is randomly selected during training.
We train the model on each of the above sequences with an equal probability of $50$\%. Both spatial and temporal training sequences are randomly sampled from the entire spatial (temporal) candidates.
During training, we randomly drop all conditions (including image latents, skeleton latents, and Plücker coordinates) with a probability of $10$\% to enable classifier-free guidance.
Please refer to the supplementary materials for more details.

\begin{table*}
\small
\caption{\textbf{Quantitative comparison on DNA-Rednering~\cite{2023dnarendering} and ActorsHQ~\cite{isik2023humanrf}.} \sysname{} surpasses baseline approaches across different settings and metrics. Note that CAT4D$^\dagger$ is our reproduced version.}
\centering

\begin{tabular}{ll|ccc|ccc}
\toprule
\multicolumn{2}{l}{\textbf{Dataset}} & \multicolumn{3}{c}{\textbf{4-View Video}} & \multicolumn{3}{c}{\textbf{8-View Video}} \\
\multicolumn{1}{l}{Method} & \multicolumn{1}{l}{Type} & PSNR$\uparrow$ & SSIM$\uparrow$ & {LPIPS$\downarrow$} & PSNR$\uparrow$ & SSIM$\uparrow$  & \multicolumn{1}{c}{LPIPS$\downarrow$} \\

\midrule

\multicolumn{2}{l}{\textbf{DNA-Rendering}~\cite{2023dnarendering}} \\
LongVolcap (4DGS)~\cite{xu2024longvolcap} & Optimization & 20.064 & 0.740 & 0.296 & 24.211 & 0.840 & 0.221 \\
GauHuman~\cite{hu2023gauhuman} & SMPL-Prior  & 18.406 & 0.723 & 0.327 & 18.818 & 0.737 & 0.316 \\
GPS-Gaussian~\cite{zheng2024gpsgaussian} & Feed-Forward & 11.250 & 0.457 & 0.460 & 17.604 & 0.714 & 0.270 \\
CAT4D$^\dagger$~\cite{wu2024cat4d} & Generation & 21.445 & 0.806 & 0.234 & 22.531 & 0.824 & 0.221 \\
Ours & Generation & \textbf{25.393} & \textbf{0.864} & \textbf{0.161} & \textbf{26.324} & \textbf{0.881} & \textbf{0.150} \\

\midrule

\multicolumn{2}{l}{\textbf{ActorsHQ}~\cite{isik2023humanrf}} \\
LongVolcap (4DGS)~\cite{xu2024longvolcap} & Optimization & 21.313 & 0.761 & 0.271 & 28.120 & 0.896 & 0.156 \\
GauHuman~\cite{hu2023gauhuman} & SMPL-Prior  & 20.449 & 0.776 & 0.275 & 21.454 & 0.803 & 0.252 \\
GPS-Gaussian~\cite{zheng2024gpsgaussian} & Feed-Forward  & 10.562 & 0.453 & 0.481 & 14.208 & 0.601 & 0.379 \\
CAT4D$^\dagger$~\cite{wu2024cat4d} & Generation & 21.562 & 0.808 & 0.229 & 23.002 & 0.873 & 0.206 \\
Ours & Generation & \textbf{27.875} & \textbf{0.903} & \textbf{0.121} & \textbf{28.747} & \textbf{0.916} & \textbf{0.110} \\

\bottomrule
\end{tabular}

\vspace{-2mm}
\label{tab:quantitative}
\end{table*}

\begin{table}
\small
\caption{\textbf{Quantitative comparison of denoising strategies.}}
\centering
\begin{tabular}{l|ccc}
\toprule

\multicolumn{1}{l}{Sampling Method} & PSNR$\uparrow$ & SSIM$\uparrow$ & \multicolumn{1}{c}{LPIPS$\downarrow$} \\

\midrule

Multi-group & 20.913 & 0.753 & 0.224 \\
Median filtering & 21.609 & 0.766 & 0.226 \\
Sliding iterative & \textbf{22.363} & \textbf{0.778} & \textbf{0.196} \\

\bottomrule
\end{tabular}

\vspace{1.2mm}
\label{tab:denoising}
\end{table}

\begin{table}
\small
\caption{\textbf{Quantitative comparison of conditioning schemes.}}
\centering
\begin{tabular}{l|ccc}
\toprule

\multicolumn{1}{c}{Condition} & PSNR$\uparrow$ & SSIM$\uparrow$ & \multicolumn{1}{c}{LPIPS$\downarrow$} \\

\midrule

w/o skeleton & 16.864 & 0.608 & 0.272 \\
w/o Plücker & 20.753 & 0.638 & 0.203 \\
Ours & \textbf{22.120} & \textbf{0.707} & \textbf{0.184} \\

\bottomrule
\end{tabular}

\vspace{-2mm}
\label{tab:conditioning}
\end{table}

\subsection{Datasets and Baselines}

\PAR{Datasets.}
We train our model on the DNA-Rendering~\cite{2023dnarendering} dataset, which contains over $2,000$ sequences of human performances in diverse outfits and dynamic motions. 
For training, we first filter out actors interacting with complex objects and then select $1,000$ sequences, each with $48$ views and $225$ frames per view, totaling $10$ million images. 
We use 16 sequences from the test set covering a variety of clothing types and action categories for quantitative comparison.
Additionally, we evaluate our model on the ActorsHQ~\cite{isik2023humanrf} dataset, which consists of $12$ sequences of human performances, to assess zero-shot generalization.

\PAR{Baselines.}
We compare our approach with multiple categories of state-of-the-art methods: the optimization-based method LongVolcap~\cite{xu2024longvolcap}, the SMPL-based method GauHuman~\cite{hu2023gauhuman}, the feed-forward method GPS-Gaussian~\cite{zheng2024gpsgaussian}, and the generation-based method CAT4D$^\dagger$~\cite{wu2024cat4d}. $\dagger$ indicates our reproduced version.

Since neither the DNA-Rendering nor ActorsHQ provides SMPL models for our test sequences, we use EasyMocap~\cite{easymocap} to extract SMPL models for test sequences, which serve as the inputs for GauHuman~\cite{hu2023gauhuman}.
For GPS-Gaussian~\cite{zheng2024gpsgaussian}, we follow their view-selection strategy: select the two input views closest to the target view as input to generate each view.
We trained CAT4D$^\dagger$~\cite{wu2024cat4d} on the processed DNA-Rendering dataset under the same training settings as our approach. CAT4D$^\dagger$~\cite{wu2024cat4d} chooses conditional views from sparse-view videos to denoise each row or column, using the same sampling sequences and conditional-view selection strategy described in \cref{sec:denoising}.

\begin{figure}[t!]
    \centering
    \includegraphics[width=0.99\linewidth]{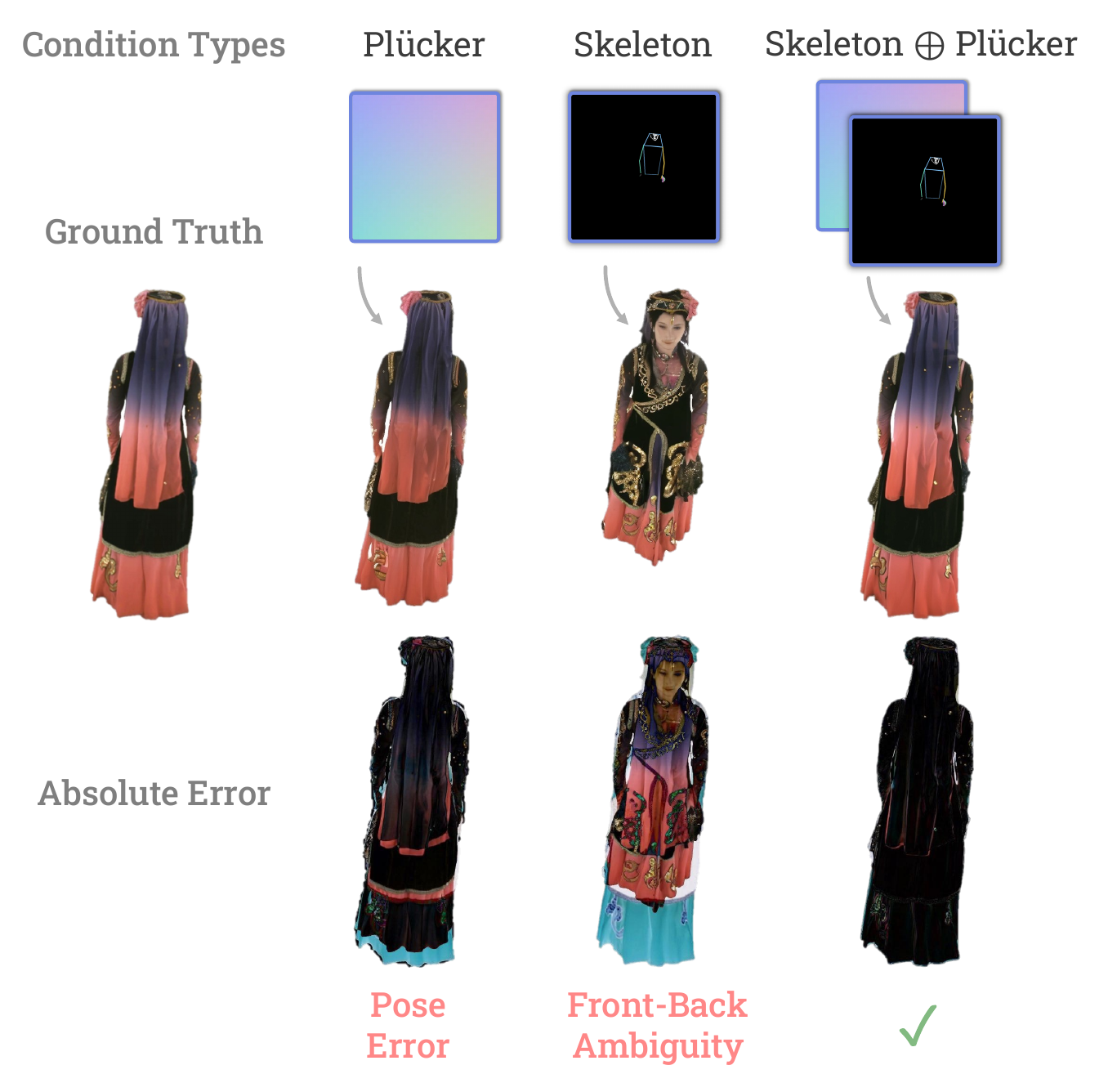}
    \vspace{-0.1in}
    \caption{\textbf{Qualitative comparison of different conditioning}. The skeleton-Plücker mixed conditioning serves as a robust human pose prior for diffusion models.}
    \vspace{-0.1in}
    \label{fig:ablation_conditioning}
\end{figure}

\subsection{Comparison to Baselines}
\label{sec:comparisons}

\begin{figure*}[t!]
    \centering
    \begin{subfigure}[b]{0.495\linewidth}
        \centering
        \includegraphics[width=\linewidth]{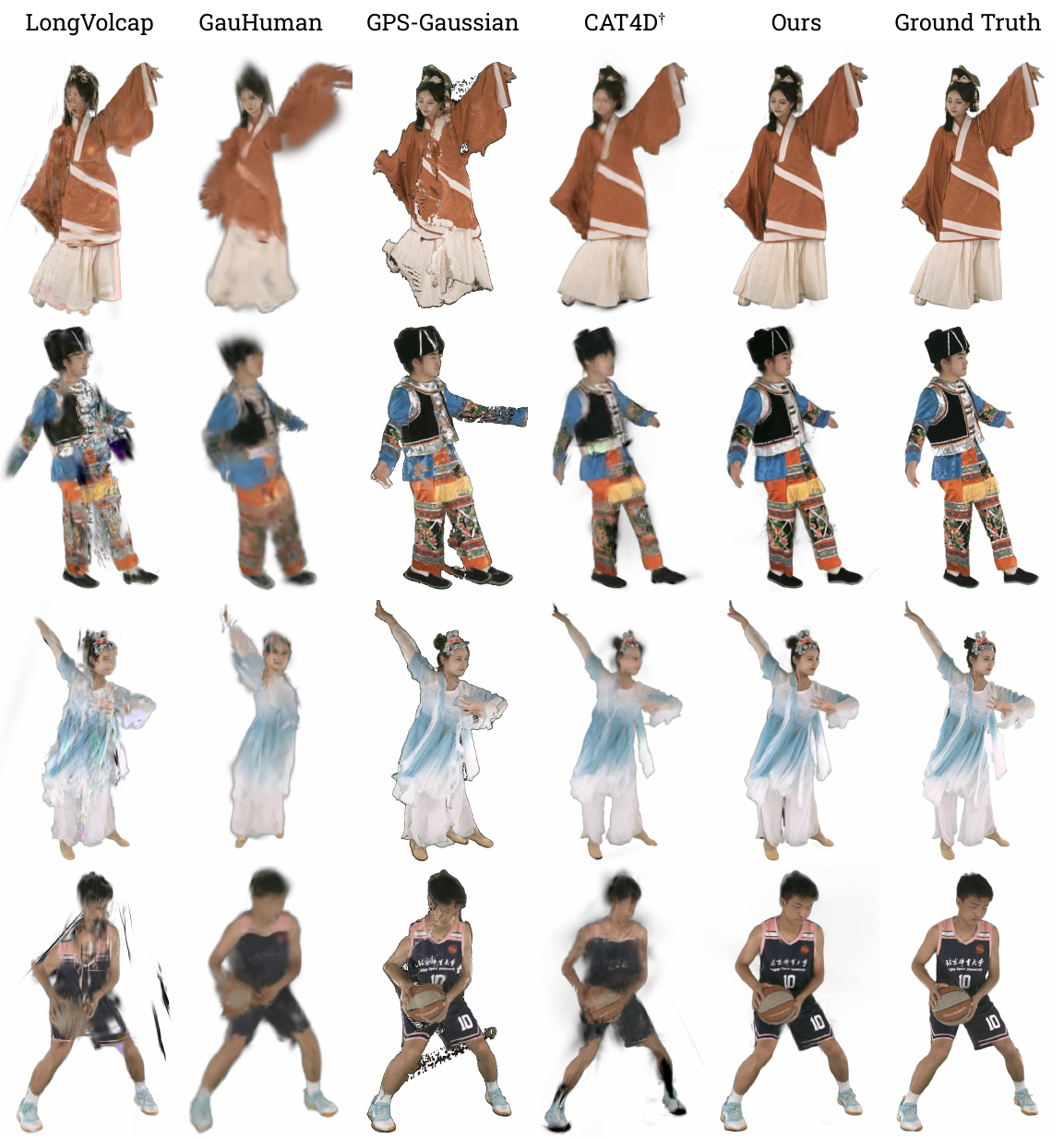}
        \caption{\textbf{Results on DNA-Rendering~\cite{2023dnarendering} test set.}}
        \label{fig:comparison_dna}
    \end{subfigure}
    \hfill
    \begin{subfigure}[b]{0.495\linewidth}
        \centering
        \includegraphics[width=\linewidth]{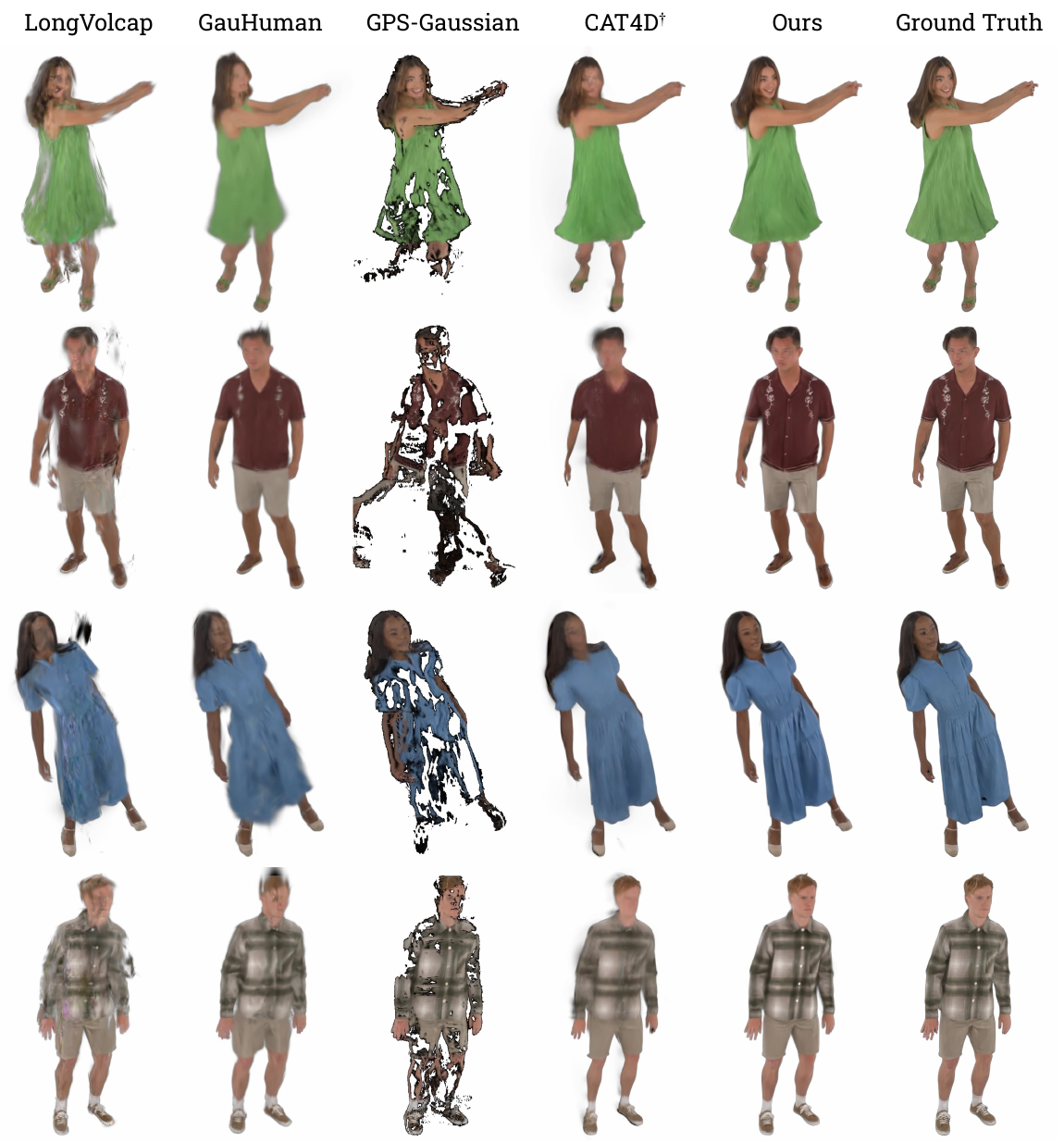}
        \caption{\textbf{Zero-shot generalization on ActorsHQ~\cite{isik2023humanrf}.}}
        \label{fig:comparison_actorshq}
    \end{subfigure}
    \vspace{-0.1in}
    \caption{\textbf{Qualitative comparison on DNA-Rendering~\cite{2023dnarendering} and ActorsHQ~\cite{isik2023humanrf}.} GPS-Gaussian uses 8 input views while all other methods use 4 input views. CAT4D$^\dagger$ is our reproduced version. Our \sysname{} consistently outperforms baselines with higher visual quality and better spatio-temporal consistency.}
    \vspace{-0.1in}
    \label{fig:comparison}
\end{figure*}

\PAR{Comparison on the DNA-Rendering~\cite{2023dnarendering} dataset.}
We provide quantitative and qualitative comparisons on the DNA-Rendering~\cite{2023dnarendering} dataset in \cref{tab:quantitative} and \cref{fig:comparison} respectively.
As shown in the visualization and metrics, our method consistently outperforms baselines with higher visual quality and better spatio-temporal consistency.
The optimization-based method LongVolcap~\cite{xu2024longvolcap} struggles with the ill-posed nature of sparse-view reconstruction, resulting in noisy renderings.
GauHuman~\cite{hu2023gauhuman} fails to reliably reconstruct performers wearing complex clothing and executing dynamic motions.
The depth estimator of GPS-Gaussian~\cite{zheng2024gpsgaussian} breaks down under our sparse-view settings, yielding fragmented results on highly dynamic sequences.
In contrast, our method not only effectively handles the challenging sparse-view setting by producing reasonable guidance from the diffusion prior, but can also generalize well to the complex motions and appearances of dynamic human performers.
Note that even with only 4 input views, our method achieves visual quality comparable to the dense reconstruction from 48 views using LongVolcap \cite{xu2024longvolcap}.

\begin{figure*}[ht!]
    \centering
    \includegraphics[width=0.99\linewidth]{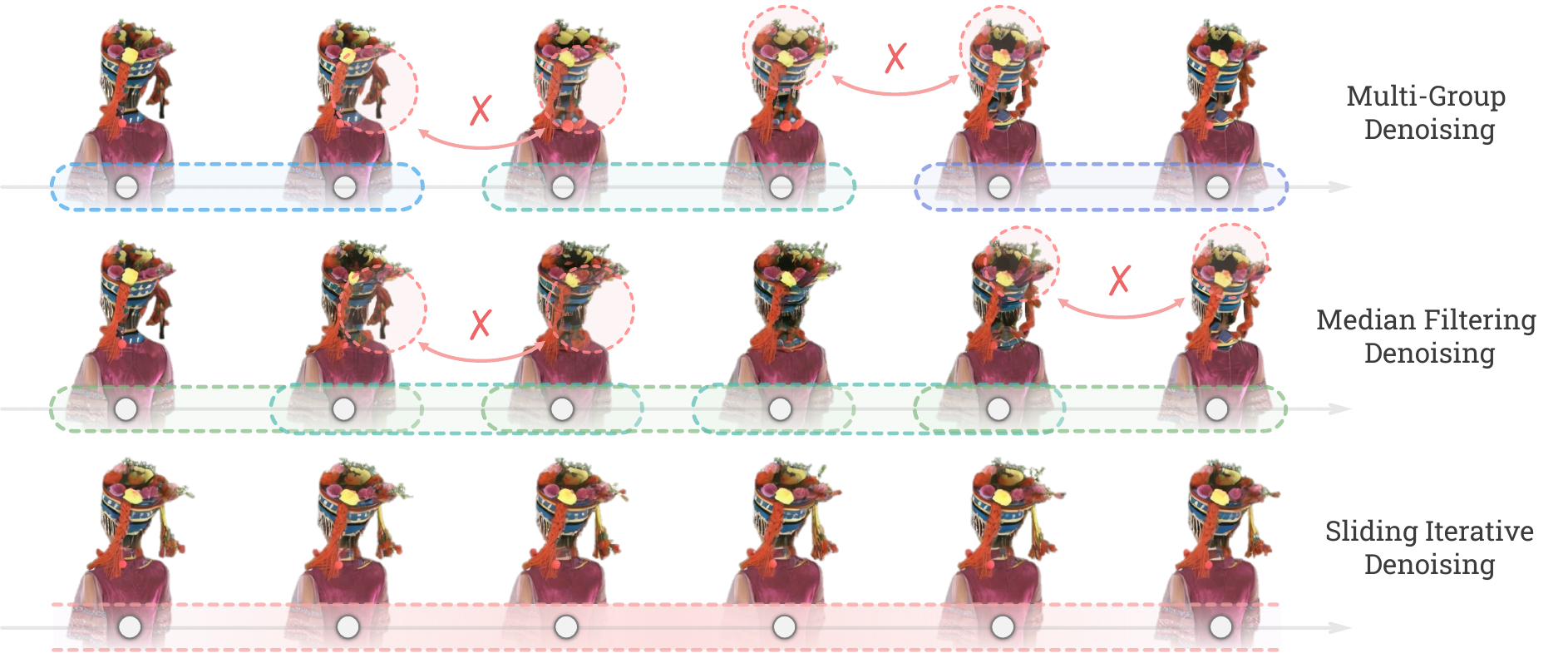}
    \vspace{-0.1in}
    \caption{\textbf{Qualitative comparison of different denoising strategies}. Our sliding iterative denoising method ensures a consistent appearance throughout a long image sequence.}
    \vspace{-0.1in}
    \label{fig:ablation_denoising}
\end{figure*}

\PAR{Comparison on the ActorsHQ~\cite{isik2023humanrf} dataset.}
As shown in \cref{tab:quantitative} and \cref{fig:comparison}, our model generalizes well to the unseen actor appearances and motions of the ActorsHQ dataset.
In comparison, the baseline methods struggle to produce coherent geometry and appearance, consistent with their results on the DNA-Rendering dataset.
Thanks to our unique model design, even with limited observations on the ActorsHQ dataset, \sysname{} consistently produces much sharper and also spatio-temporally consistent human performance reconstruction results.

\subsection{Ablation Study}
\label{sec:ablation}

We perform two ablation studies on the DNA-Rendering dataset~\cite{2023dnarendering}, using three challenging motion sequences to evaluate the sliding iterative denoising mechanism and six complex-clothing sequences to assess the skeleton–Plücker conditioning scheme.

\PAR{Diffusion denoising strategy.}
We compare three diffusion sampling strategies: the multi-group denoising approach that divides the data into multiple groups, the median filtering approach that takes the median values of the denoised results of overlapping images between groups, and our sliding iterative diffusion denoising strategy.
As shown in \cref{fig:ablation_denoising} and \cref{tab:denoising}, the multi-group denoising approach does not take the temporal and spatial correlation between groups into consideration, leading to sudden jumps in the generated results between different segments.
The median filtering approach slightly mitigates the intra-group inconsistency of the multi-group approach by taking the median values of the denoised results between different windows.
However, the computational cost of this approach is inversely proportional to the overlap ratio, and could still produce inconsistency if the window doesn't overlap enough.
In contrast, our sliding iterative denoising strategy introduces a smoothness-inductive bias during the denoising process of the diffusion model, at the same time maintaining constant computational cost by merging the sliding operation with the denoising steps. This process produces more consistent and globally accurate results compared to the two solutions.

\PAR{Conditioning scheme.}
As shown in \cref{fig:ablation_conditioning} and \cref{tab:conditioning}, we compare three conditioning schemes: w/o skeleton, w/o Plücker, and our skeleton-Plücker.
The w/o-skeleton conditioning has limited camera control over the generated content, producing large misalignment due to the ill-posed nature of the generative problem.
The w/o-Plücker conditioning can provide fine-grain control over the generated human, but it may struggle to disambiguate between the front and back, left and right of the generation target, producing inconsistent guidance for the reconstruction module.
In comparison, our conditioning scheme combines the merits of the camera and pose control signal of the Plücker and skeleton embedding, generating consistent and controllable novel view results of the target human actor.

\section{Conclusion}
\label{sec:conclusion}

This paper introduces \sysname{}, a novel diffusion model capable of generating high-resolution (1024p) and 4D-consistent human images from sparse-view inputs. We propose a novel sliding iterative denoising strategy to enhance spatial and temporal consistency while maintaining high computational efficiency. To further improve motion accuracy and visual quality, we introduce a 4D pose conditioning mechanism that leverages human skeletons. Our method demonstrates superior performance in capturing fine details and complex human motions from sparse inputs compared to existing state-of-the-art approaches.

Despite promising results, our method still faces certain limitations. First, higher-resolution (4K) videos are not supported due to constraints in base models. Second, our method may struggle with scenarios involving complex human-object interactions. Finally, our current method cannot achieve novel-pose rendering because it requires input videos to constrain the spatial consistency of the generated videos. Addressing these challenges represents interesting directions for future research.

\PAR{Acknowledgments.}
This work was partially supported by the National Key R\&D Program of China (No.2024YFB2809105), NSFC (No.U24B20154, No.\-62172364), Zhejiang Provincial Natural Science Foundation of China (No.LR25F020003), Ant Research, and Information Technology Center and State Key Lab of CAD\&CG, Zhejiang University.

{
    \small
    \bibliographystyle{ieeenat_fullname}
    \bibliography{main}
}

\clearpage
\setcounter{page}{1}
\maketitlesupplementary
\appendix

\section{Model Details}
\label{sec:sup_model}

\PAR{Training.}
We initialize our model from Stable Diffusion 2.1~\cite{Rombach_2022_CVPR} developed by Diffusers~\cite{von-platen-etal-2022-diffusers}. 
We apply full-parameter fine-tuning to the diffusion model for $200$k iterations with a batch size of $32$ and a learning rate of $10^{-5}$ using 32 NVIDIA H20 GPUs.
To accommodate input images with the conditions, we expand the input channels of the model's first convolutional layer from 4 to 15, consisting of 4 channels for image latents, 4 for skeleton latents, 6 for Plücker embeddings, and 1 for a conditional mask. 
Following the previous work \cite{gao2024cat3d}, the conditional mask is a binary indicator specifying whether an image serves as a conditioning input or a target.

\PAR{Sampling.}
Following Stable Diffusion 2.1~\cite{Rombach_2022_CVPR,von-platen-etal-2022-diffusers}, we use DPM-Solver++~\cite{lu2022dpm} with $24$ sampling steps and a classifier-free guidance scale of $3.0$. Our sliding iterative denoising strategy takes approximately $2$ minutes to generate a sample sequence of length $48$ when executed on a single A100 GPU. To improve efficiency, we parallelize the denoising process across $8$ A100 GPUs.

\PAR{4D reconstrcution.}
We employ LongVolcap~\cite{xu2024longvolcap} to reconstruct the 4D human performances from the generated multi-view videos. LongVolcap is an enhanced version of 4DGS~\cite{yang2023gs4d} with the ability of effectively reconstructing long volumetric videos with a temporal Gaussian hierarchy representation. We initialize the 4D Gaussian primitives with the coarse geometry obtained using the predicted foreground masks and the space carving algorithm~\cite{kutulakos2000theory,xu20244k4d}. We then follow the same training and evaluation settings as in the original paper~\cite{xu2024longvolcap} to reconstruct the 4D human performances. Specifically, we optimize the model with the Adam optimizer~\cite{kingma2014adam} with a learning rate of $1.6 e^{-4}$, each model is trained for $100$k iterations for a sequence of $7200$ frames, which takes around $1$ hour on a single NVIDIA RTX 4090 GPU.

\section{Datasets Details}
\label{sec:sup_dataset}

We conduct extensive processing on the original DNA-Rendering~\cite{2023dnarendering} dataset to generate high-quality multi-view videos along with additional masks and skeletons for training and evaluation. The processing pipeline includes camera re-calibration, color correction matrices (CCMs) optimization, foreground mask prediction, and human skeleton estimation. We provide detailed descriptions of each step below.

\PAR{Camera calibration.}
We empirically found that the camera parameters provided in the DNA-Rendering dataset are not accurate enough for reconstruction verified with 3D Gaussian Splatting (3DGS)~\cite{kerbl3Dgaussians}. In order to achieve pixel-level accuracy, we first re-calibrated the camera parameters using Colmap~\cite{schoenberger2016sfm,schoenberger2016mvs}. We then optimized the color correction matrix for each camera to ensure consistent color across different views.

\begin{figure}[ht!]
    \centering
    \includegraphics[width=1.0\linewidth]{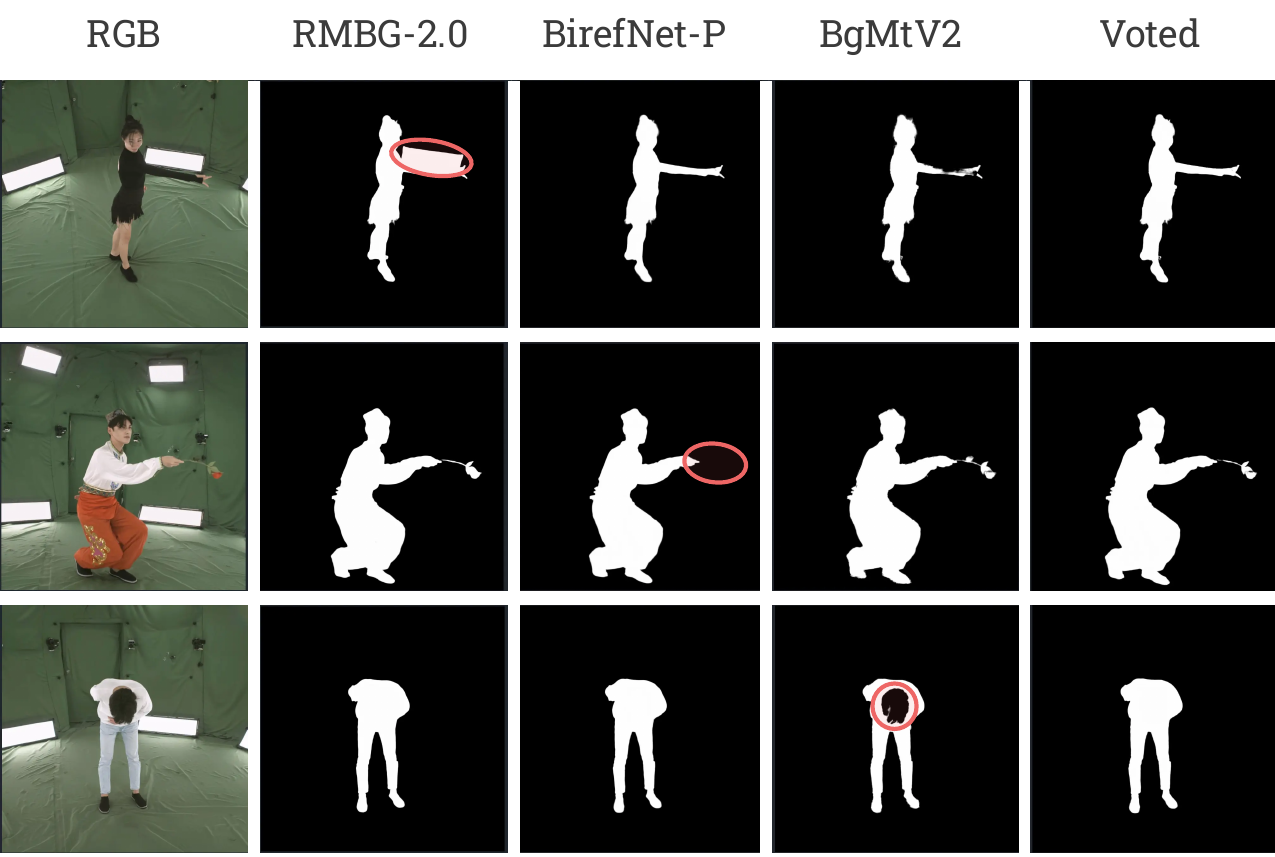}
    \vspace{-0.1in}
    \caption{Our voting strategy effectively leverages the strengths of different background removal methods to produce robust foreground masks.}
    \vspace{-0.1in}
    \label{fig:supp_fmask}
\end{figure}

\PAR{Foreground mask prediction.}
There are only a few (around 1/6) sequences in the DNA-Rendering dataset that provide ground truth foreground masks. To obtain accurate foreground masks, we leverage three state-of-the-art background removal methods, namely RMBG-2.0~\cite{zheng2024birefnet}, BiRefNet-Portrait~\cite{zheng2024birefnet}, and BackgroundMattingV2~\cite{lin2021real}, and combine their predictions using a voting mechanism to fully leverage the strengths of each approach. Specifically, we found that RMBG-2.0 may incorrectly recognize background objects as foreground, BiRefNet-Portrait may segment small objects as background, and BackgroundMattingV2 may produce inaccurate results for certain human poses. We demonstrate the effectiveness of the voting strategy in \cref{fig:supp_fmask}.

\begin{figure*}
    \centering
    \includegraphics[width=1.0\linewidth]{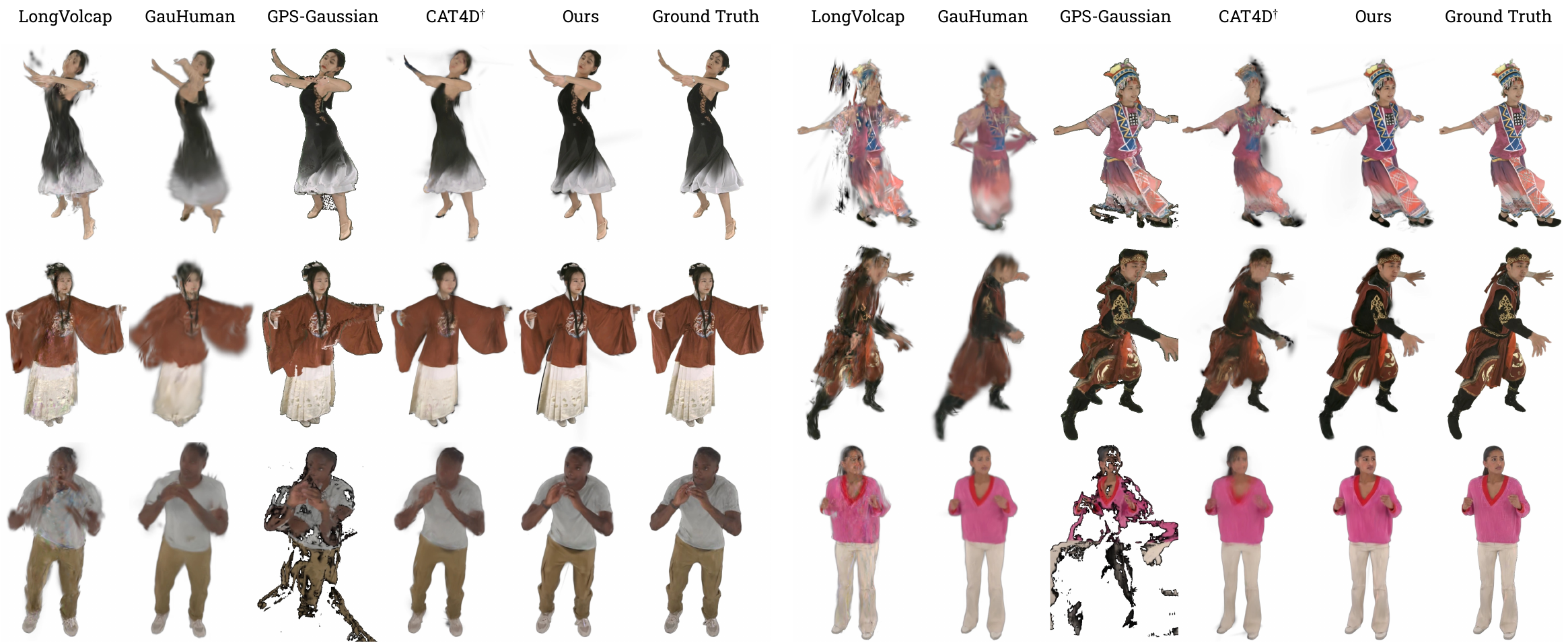}
    \vspace{-0.1in}
    \caption{\textbf{More qualitative comparisons on DNA-Rendering~\cite{2023dnarendering} and ActorsHQ~\cite{isik2023humanrf}.} GPS-Gaussian uses 8 input views while all others use 4 input views, and CAT4D$^\dagger$ is our reproduced version. Our \sysname{} consistently outperforms baselines with higher visual quality and better spatio-temporal consistency.}
    \vspace{-0.1in}
    \label{fig:supp_comparison}
\end{figure*}

\PAR{Human skeleton estimation.}
Similar to the foreground mask, only a few sequences have ground truth human skeletons. We thus adopt the state-of-the-art human skeleton estimation model, Sapiens~\cite{khirodkar2024sapiens}, to predict the 2D human skeleton for each frame. We additionally adjust the transparency of the skeleton colors based on the confidence scores of the skeleton, which helps to visualize and encode the uncertainty of the skeleton estimation. After obtaining the 2D human skeletons, we then triangulate them to obtain the 3D human skeleton sequence, which can be further used for projection and evaluation.

\begin{figure}[ht!]
    \centering
    \includegraphics[width=1.0\linewidth]{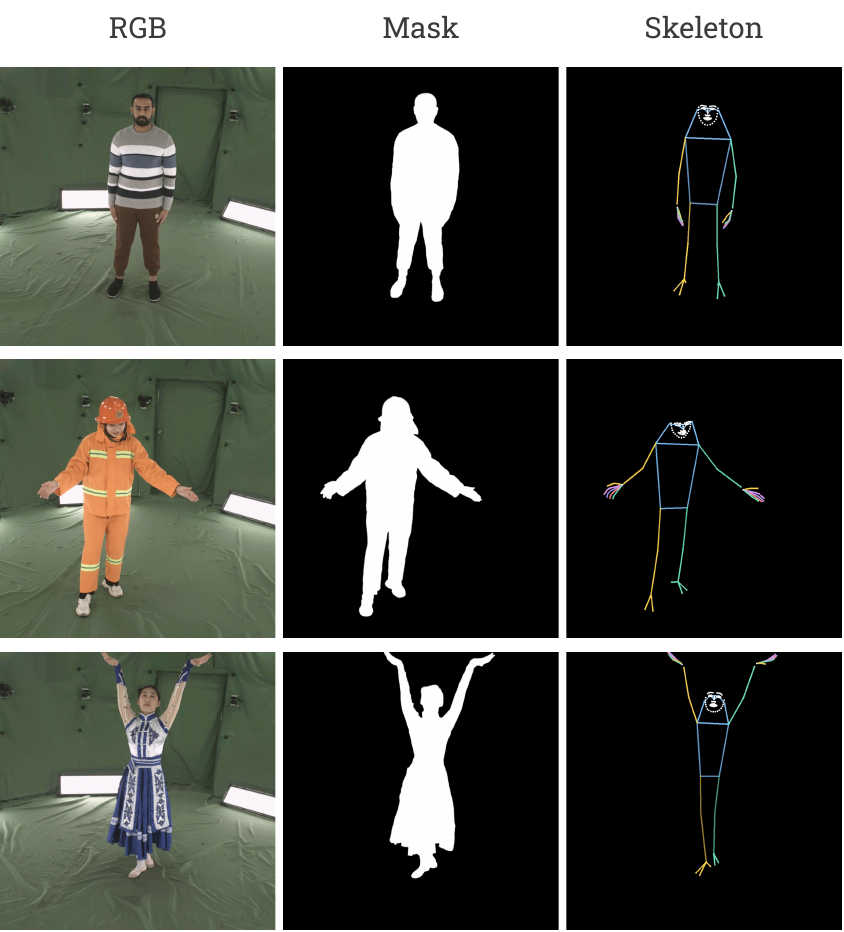}
    \vspace{-0.1in}
    \caption{High-quality foreground masks and human skeletons predicted using state-of-the-art methods.}
    \vspace{-0.1in}
    \label{fig:supp_data_sample}
\end{figure}

We demonstrate the processed data samples in \cref{fig:supp_data_sample}.
We plan to release the additionally processed data under the DNA-Rendering open-source license to facilitate future research within the community.

\PAR{Dataset filtering.}
DNA-Rendering~\cite{2023dnarendering} contains many scenes involving human-object interactions, such as writing on a desk, playing guitar, or organizing items. Since the diversity of objects is significantly greater than that of humans, training generative models typically requires extensive object datasets (e.g., Objaverse~\cite{objaverse,objaverseXL}). To address the relatively limited scale of the DNA-Rendering dataset, we employed the Llama Vision 3.2 model to classify all scenes and filtered out those containing large objects to avoid potential model collapse during training.

Nevertheless, we observe that even though the training dataset does not include objects, our model successfully generalizes to scenes featuring simple objects, such as the basketball player shown in \cref{fig:supp_gen_recon}.

\section{Additional Comparisons}
\label{sec:sup_comparison}

\begin{figure*}[ht!]
    \centering
    \includegraphics[width=1.0\linewidth]{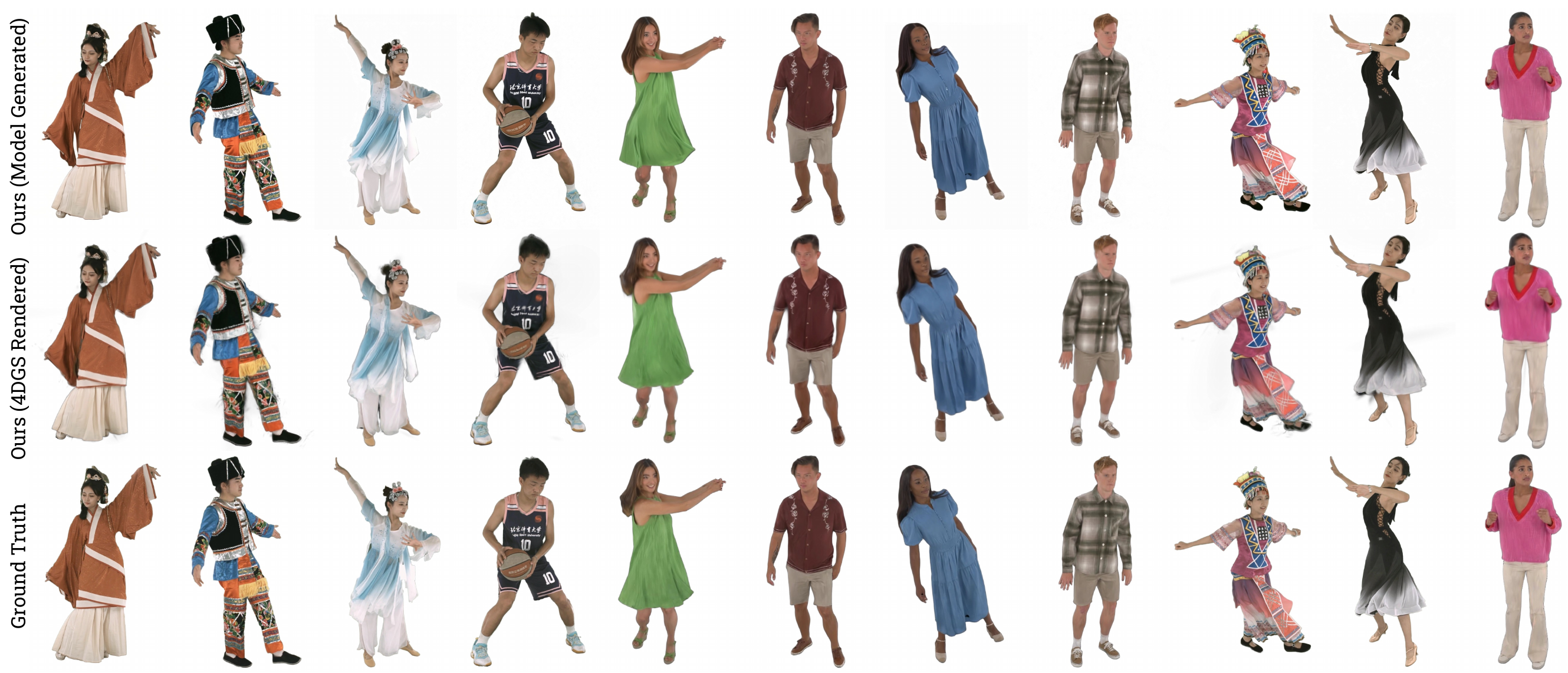}
    \vspace{-0.1in}
    \caption{Qualitative comparisons between novel views generated by our model and those rendered from the 4DGS model reconstructed using LongVolcap~\cite{xu2024longvolcap}.}
    \vspace{-0.1in}
    \label{fig:supp_gen_recon}
\end{figure*}

\PAR{More qualitative results.}
We provide additional comparisons with baselines in \cref{fig:supp_comparison}. Results show that our method consistently outperforms the baselines in terms of visual quality and fine details.

\PAR{Diffusion generation vs. 4DGS rendering.}
Although our model already supports novel-view synthesis, we choose to optimize a 4DGS model using LongVolcap~\cite{xu2024longvolcap} to enable real-time rendering. As shown in \cref{fig:supp_gen_recon}, our model can generate high-fidelity human videos, but they still inevitably exhibit spatio-temporal inconsistencies. Reconstructing a 4DGS model further alleviates these inconsistencies, though at the cost of reduced sharpness compared to the originally generated images.


\end{document}